\documentclass{article}

\usepackage{microtype}
\usepackage{graphicx}
\usepackage{subcaption}
\usepackage{booktabs} % for professional tables

\usepackage{hyperref}
\mathchardef\mhyphen="2D % Define a "math hyphen"

\usepackage[accepted]{icml2023}

\usepackage{amsmath}
\usepackage{amssymb}
\usepackage{mathtools}
\usepackage{braket}
\usepackage{amsthm}
\usepackage{physics}
\usepackage{xspace}
\usepackage{xcolor}
\usepackage{enumitem}
\usepackage{multirow}
\usepackage{tikz}
\usepackage{makecell}
\usepackage[textsize=tiny]{todonotes}

\usepackage[capitalize,noabbrev]{cleveref}

\theoremstyle{plain}

\theoremstyle{definition}

\theoremstyle{remark}

\newcommand{\ie}{\textit{i.e.}\xspace}

\newcommand{\img}[1]{\includegraphics[scale=0.17]{figs/images_of_figure2/#1}}
\newcommand{\imgpat}[1]{\includegraphics[scale=0.29]{figs/PAT_images_R_LPIPS/#1}}

\icmltitlerunning{Towards Adversarially Robust Perceptual Similarity Metrics}

\begin{document}

\twocolumn[
\icmltitle{R-LPIPS: An Adversarially Robust Perceptual Similarity Metric}

% It is OKAY to include author information, even for blind
% submissions: the style file will automatically remove it for you
% unless you've provided the [accepted] option to the icml2023
% package.

% List of affiliations: The first argument should be a (short)
% identifier you will use later to specify author affiliations
% Academic affiliations should list Department, University, City, Region, Country
% Industry affiliations should list Company, City, Region, Country

% You can specify symbols, otherwise they are numbered in order.
% Ideally, you should not use this facility. Affiliations will be numbered
% in order of appearance and this is the preferred way.

\begin{icmlauthorlist}
\icmlauthor{Sara Ghazanfari}{1}
\icmlauthor{Siddharth Garg}{1}
\icmlauthor{Prashanth Krishnamurthy}{1}
\icmlauthor{Farshad Khorrami}{1}
\icmlauthor{Alexandre Araujo}{1}
\end{icmlauthorlist}

% \icmlaffiliation{1}{New York University, NY, USA}
% \icmlaffiliation{comp}{Company Name, Location, Country}
% \icmlaffiliation{sch}{School of ZZZ, Institute of WWW, Location, Country}
\icmlaffiliation{1}{Department of Electrical and Computer Engineering, New
York University, NY, USA}

\icmlcorrespondingauthor{Sara Ghazanfari}{sg7457@nyu.edu}

% You may provide any keywords that you
% find helpful for describing your paper; these are used to populate
% the "keywords" metadata in the PDF but will not be shown in the document
\icmlkeywords{Perceptual Similarity Metric, Adversarial Robustness}

\vskip 0.3in
]

% this must go after the closing bracket ] following \twocolumn[ ...

% This command actually creates the footnote in the first column
% listing the affiliations and the copyright notice.
% The command takes one argument, which is text to display at the start of the footnote.
% The \icmlEqualContribution command is standard text for equal contribution.
% Remove it (just {}) if you do not need this facility.

\printAffiliationsAndNotice{}  % leave blank if no need to mention equal contribution
% \printAffiliationsAndNotice{} % otherwise use the standard text.

\begin{abstract}
Similarity metrics have played a significant role in computer vision to capture the underlying semantics of images. In recent years, advanced similarity metrics, such as the Learned Perceptual Image Patch Similarity (LPIPS), have emerged. These metrics leverage deep features extracted from trained neural networks and have demonstrated a remarkable ability to closely align with human perception when evaluating relative image similarity. However, it is now well-known that neural networks are susceptible to adversarial examples, i.e., small perturbations invisible to humans crafted to deliberately mislead the model. Consequently, the LPIPS metric is also sensitive to such adversarial examples. This susceptibility introduces significant security concerns, especially considering the widespread adoption of LPIPS in large-scale applications. In this paper, we propose the Robust Learned Perceptual Image Patch Similarity (R-LPIPS) metric, a new metric that leverages adversarially trained deep features. Through a comprehensive set of experiments, we demonstrate the superiority of R-LPIPS compared to the classical LPIPS metric.
The code is available at \url{https://github.com/SaraGhazanfari/R-LPIPS}.
\end{abstract}

\section{Introduction}
\label{secton:introduction}

The ability to compare data points is fundamental in many areas of machine learning.
For many years, 
The $\ell_p$ distance metric, for instance, is a 
well-established mathematical tool for measuring 
differences between data points. However, in the context of computer vision, these metrics primarily focus on pixel-wise differences and fail to capture the semantic information of the images. 
This limitation becomes especially evident in high-dimensional settings, where two high-definition images depicting the same scene, \ie, sharing the same underlying informational content, are far apart in terms of $\ell_p$ distance metrics. 

Perceptual metrics~\cite{wang2003multiscale,wang2004image,hore2010image,zhang2011fsim,mantiuk2011hdr,zhang2018unreasonable} have been adopted for their ability to closely align with human perception when assessing relative image similarity.
These metrics successfully capture the underlying semantics of images, providing a more accurate reflection of human judgment.
To compute the ``distance'' between images, these metrics operate on the features of the images instead of the raw images in the image space.
For example, the Learned Perceptual Image Patch Similarity \cite{zhang2018unreasonable} (LPIPS) metric takes the Euclidean distance over the deep features (latent space) of a trained neural network.
This new semantic measure has been shown to outperform all previous metrics by large margins due to the capabilities of a neural network to learn \emph{good features}.

Notwithstanding their remarkable success of neural networks in a range of tasks, neural networks are also known to be sensitive to adversarial perturbations~\cite{goodfellow2014explaining,madry2017towards}, \ie, small perturbations invisible to humans crafted to deliberately mislead the model.  
Given that the LPIPS metric is based on the feature of a trained neural network, it should not come as a surprise that this metric is also sensitive to adversarial perturbations~\cite{kettunen2019lpips}, \ie, invisible perturbations to an image which considerably modify the LPIPS value.
This raises significant security concerns as similarity metrics are already in wide use, for instance, in detecting cases of online copyright infringement and digital forensics. 

In this paper, we propose a thorough analysis of the LPIPS metric and empirically show that this metric, which is based on the learned feature of a trained network, is sensitive to adversarial attacks.
Then, we introduce the Robust Learned Perceptual Image Patch Similarity (R-LPIPS) which leverages adversarially trained deep features and shows that is robust to adversarial perturbations.
Our contributions can be summarized as follows.
\begin{enumerate}[parsep=0pt,leftmargin=12pt,topsep=0pt]
  \item We show that the LPIPS metric is sensitive to adversarial perturbation by showing that there exist small $\ell_\infty$ perturbations such that the LPIPS between a reference image and the perturbed image is large.
  \item We propose the use of Adversarial Training~\cite{madry2017towards} to build a new Robust Learned Perceptual Image Patch Similarity (R-LPIPS) that leverages adversarially trained deep features.
  \item Based on an adversarial evaluation, we demonstrate the robustness of R-LPIPS to adversarial examples compared to the LPIPS metric.
  \item Finally, based on the work of~\citet{laidlaw2020perceptual}, we showed that the perceptual defense achieved over LPIPS metrics could easily be broken by stronger attacks developed based on R-LPIPS.
\end{enumerate}

\section{Related Work}
\label{sec:Related Papers}

In this section, we provide a comprehensive review of related works on perceptual and similarity metrics. 
Additionally, we propose a concise review of adversarial attacks and adversarial robustness.
Finally, we offer a brief summary of research that explores the intersection of robustness and similarity metrics. 

\paragraph{Similarity Metrics.}
The $\ell_2$ Euclidean distance, a classic per-pixel measure, assumes pixel-wise independence and is often used for regression problems.
However, it is insufficient for structured outputs like images due to its inability to effectively capture perceptual changes like blurring.
Peak Signal-to-Noise Ratio (PSNR) measures the quality degradation in a reconstructed or compressed image/video by calculating the ratio of peak signal power to the mean squared error.
Despite its wide usage, it does not correlate well with perceived image quality.
The Structural Similarity Index (SSIM), proposed by~\cite{wang2004image}, is a perceptual metric that quantifies the structural similarity between two images or video frames.
It takes into account luminance, contrast, and structural similarities and has been shown to correlate well with human visual perception.
SSIM calculates local measures of similarity by comparing small image patches and then computes the average similarity across the entire image.
The Feature Similarity Index for Image Quality Assessment (FSIM), proposed by~\cite{zhang2011fsim}, is a perceptual metric that evaluates image quality by quantifying the similarity between two images based on their features.
FSIM uses phase congruency for feature significance and image gradient magnitude for feature similarity, ensuring consistency across varying lighting conditions.
These two metrics are often considered to be a better indicator of perceived image quality compared to PSNR, as they correlate better with human visual perception.

More recently, the Learned Perceptual Image Patch Similarity (LPIPS) proposed by~\cite{zhang2018unreasonable} was developed and aimed at providing a more accurate measure of the perceptual similarity between two images.
Instead of comparing raw pixel data, LPIPS uses deep learning to calculate the perceptual difference between images. 
Specifically, it uses a deep convolutional neural network, pretrained on an image classification task, to extract features from the images.
Then, it computes the distance between these feature vectors to calculate the perceptual similarity.
Any neural network architecture can be used for the LPIPS metric, \cite{zhang2018unreasonable} experimented with several well-known architectures such as SqueezeNet~\cite{iandola2016squeezenet}, AlexNet~\cite{krizhevsky2017imagenet}, and VGG~\cite{simonyan2014very} and showed that the AlexNet architecture offers the best performance.

To evaluate the quality of this metric with respect to human perception compared to other perceptual metrics, \citet{zhang2018unreasonable} introduced the Berkeley-Adobe Perceptual Patch Similarity (BAPPS) dataset.
The BAPPS dataset is a large-scale, highly diverse dataset of perceptual judgments used to evaluate perceptual similarity metrics.
It contains pairs of images along with human judgments of their perceptual similarity, which serves as ground truth data.

\begin{figure*}[h]
  \centering
  \hfill \img{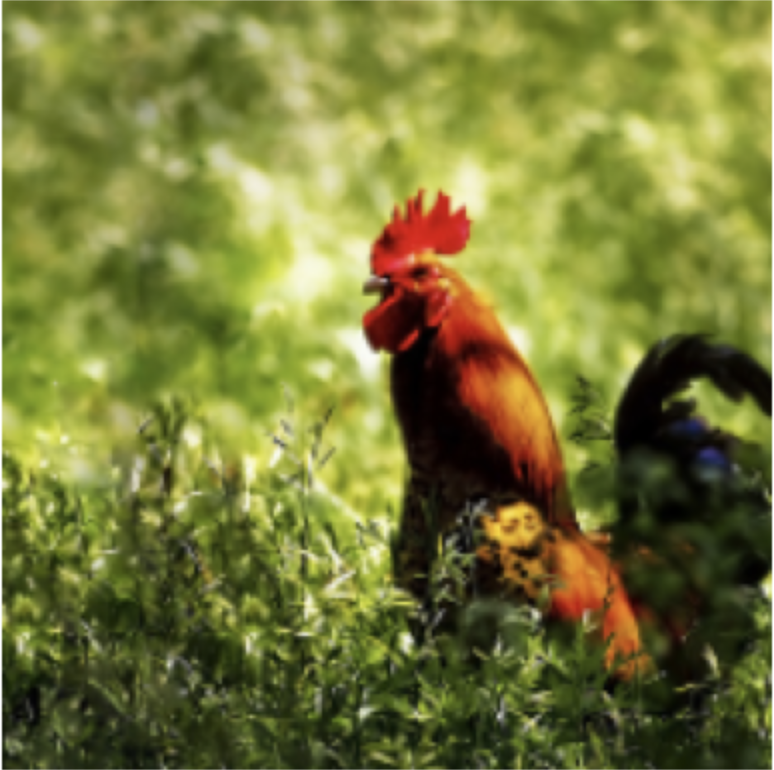} \hfill \img{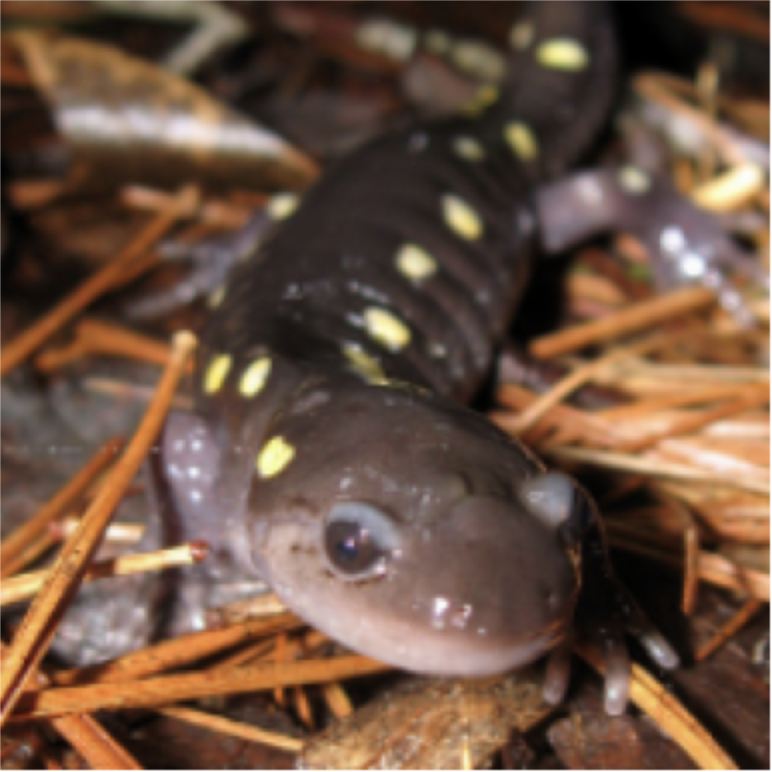} \hfill \img{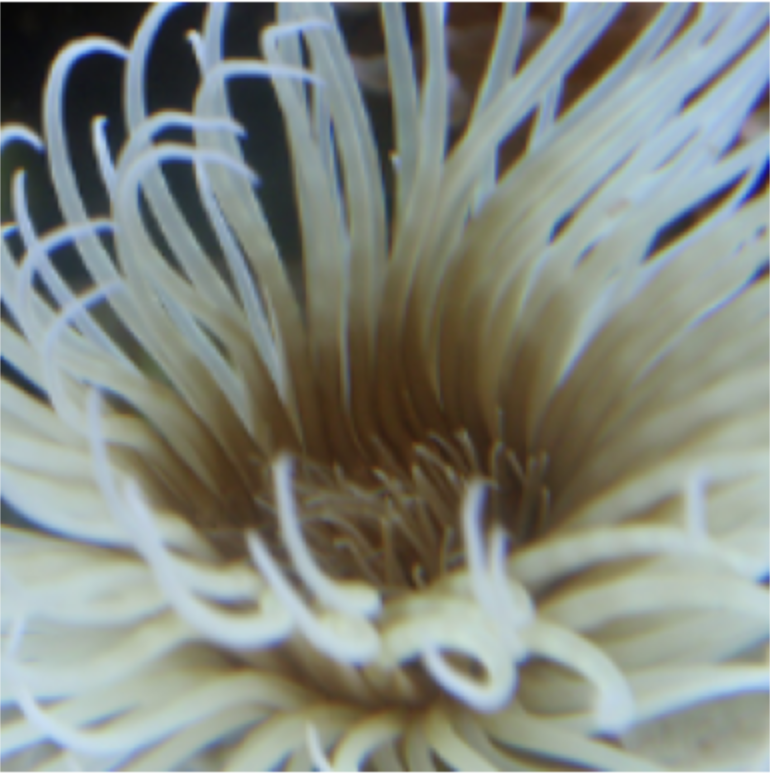} \hfill \img{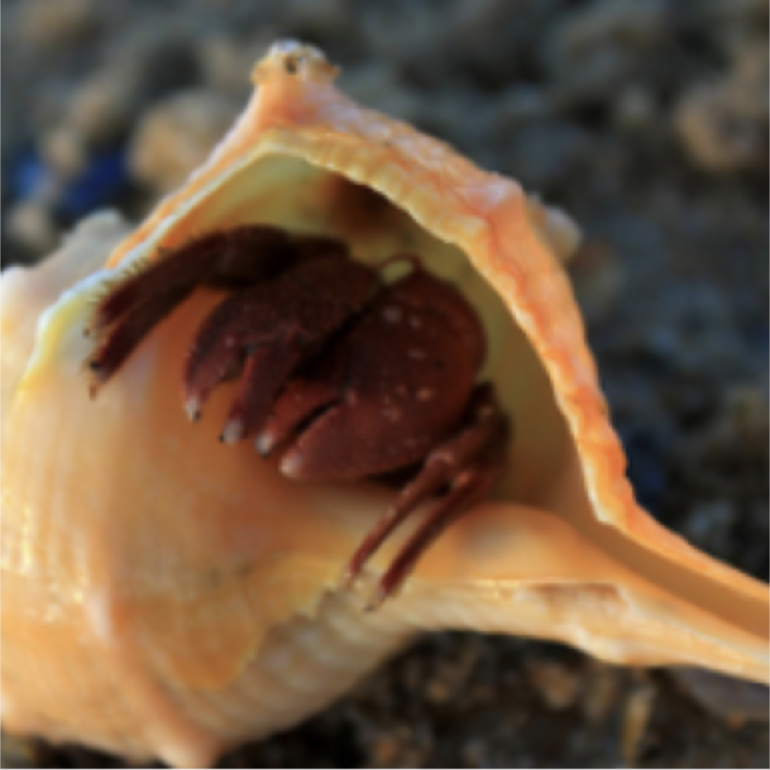} \hfill \img{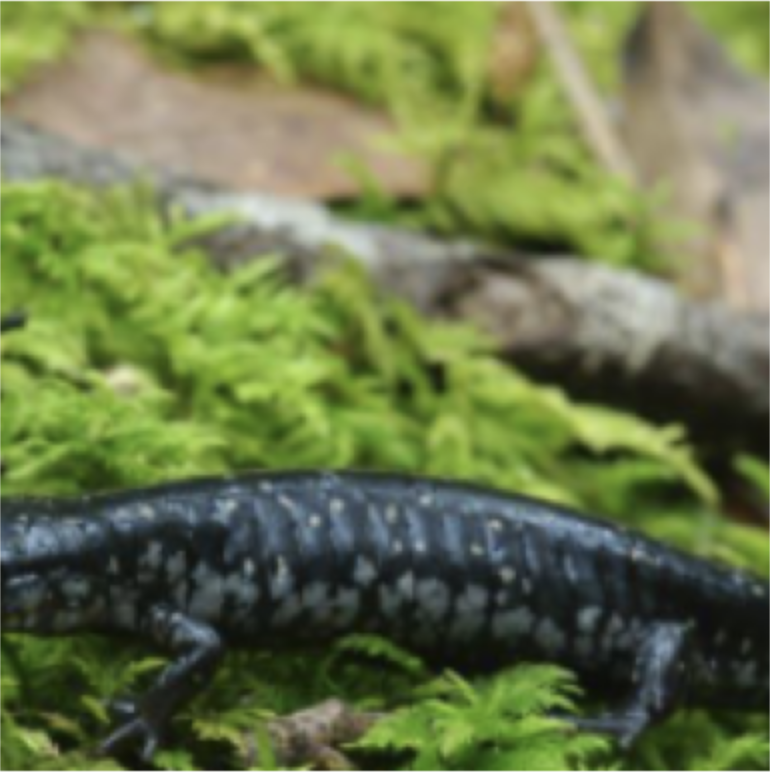} \hfill \img{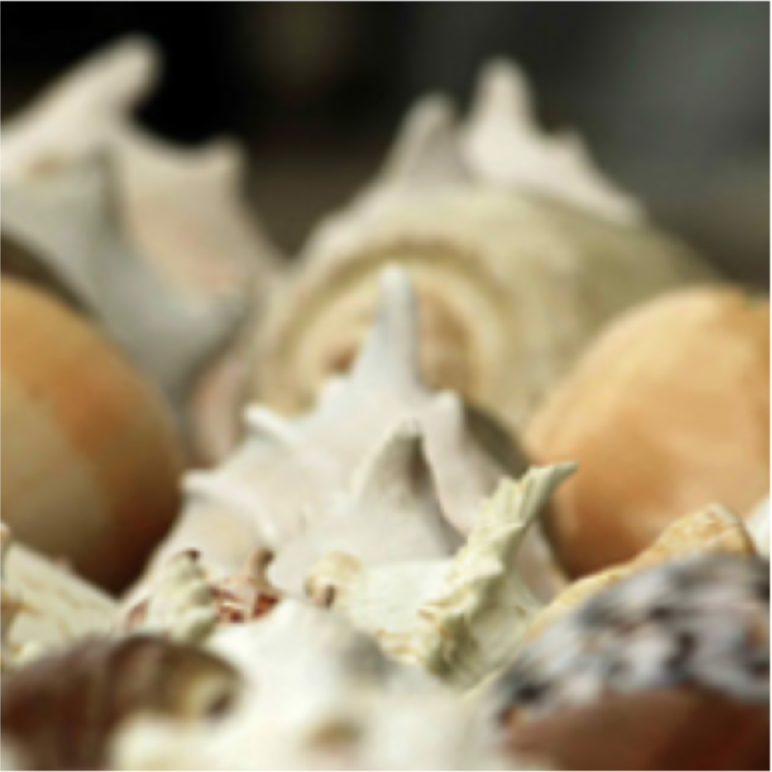} \hfill \img{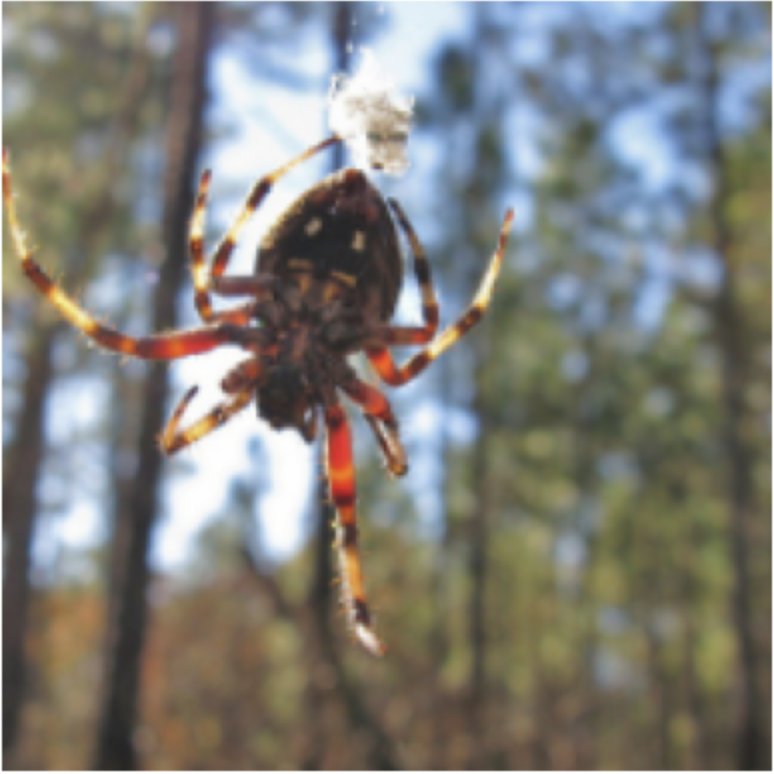} \hfill \\
  \hfill \img{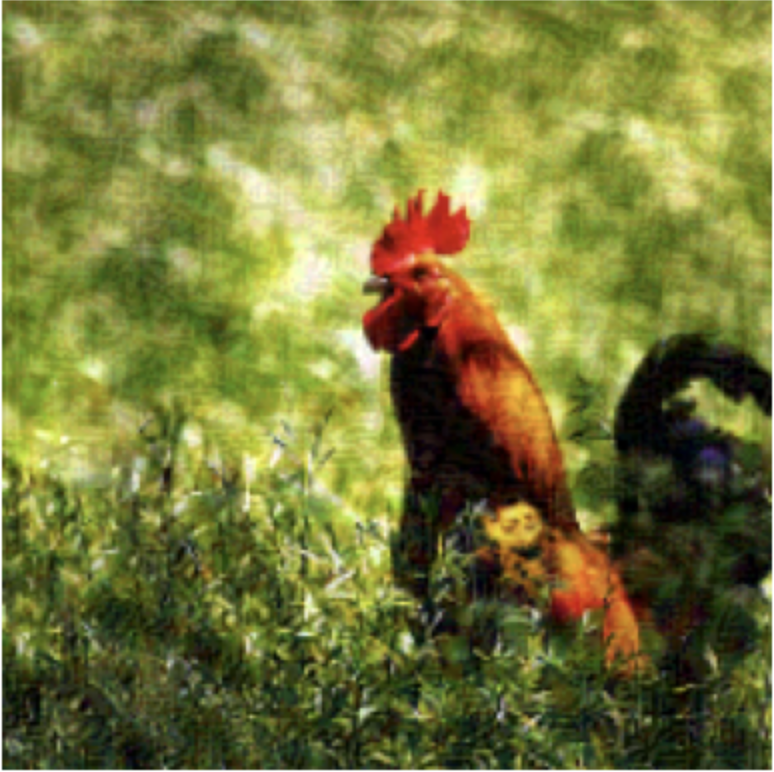} \hfill \img{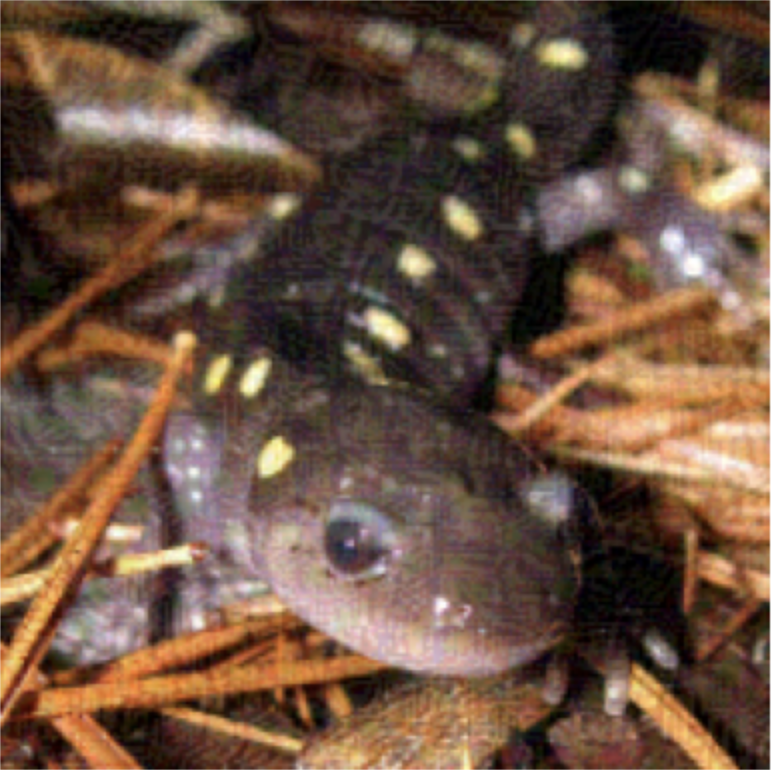} \hfill \img{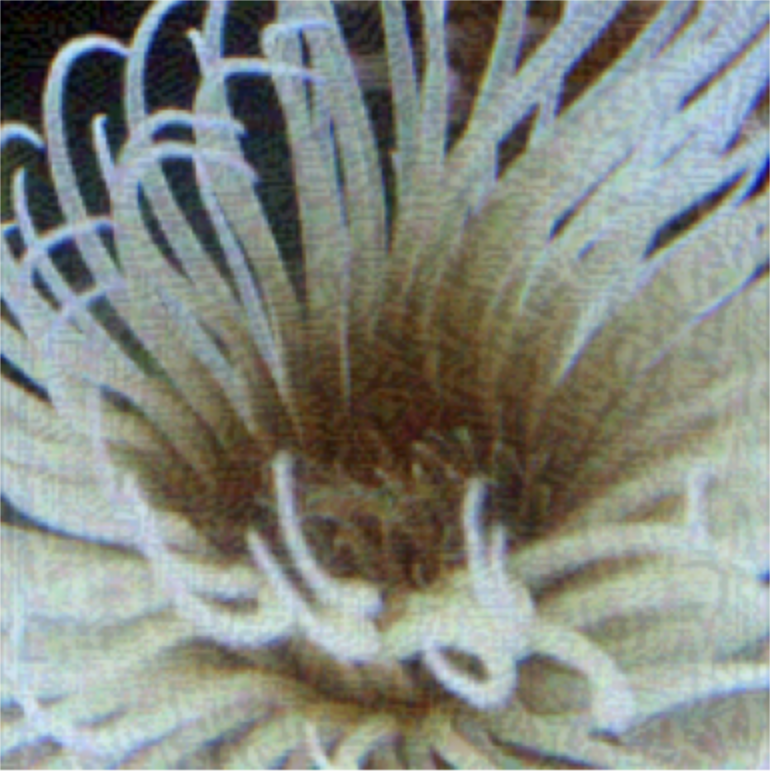} \hfill \img{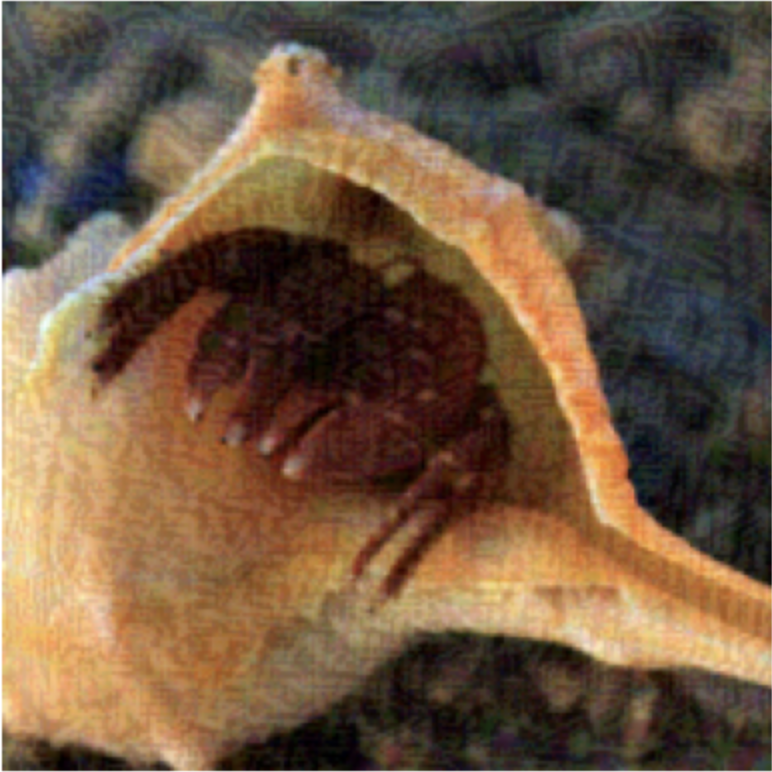} \hfill \img{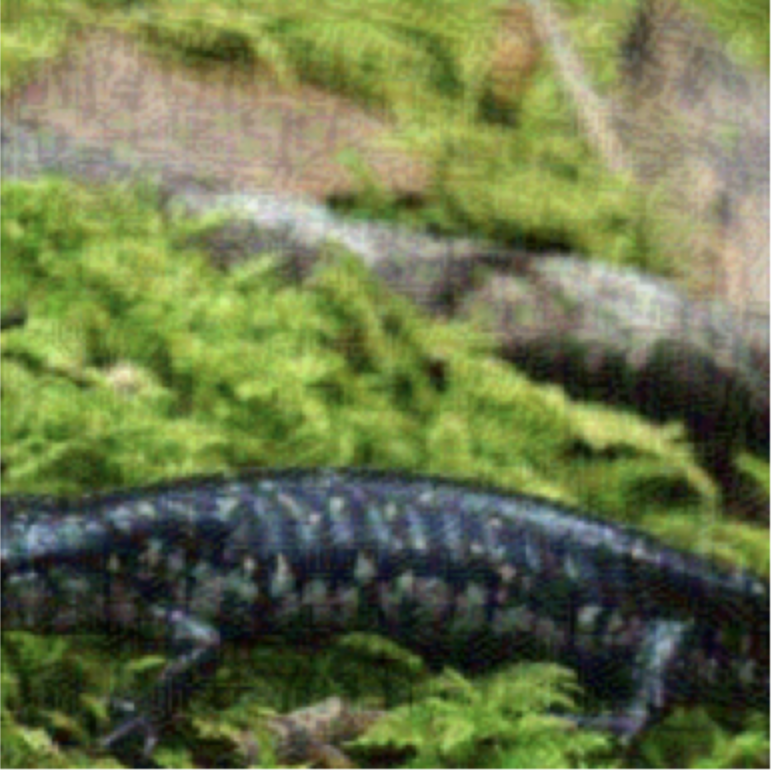} \hfill \img{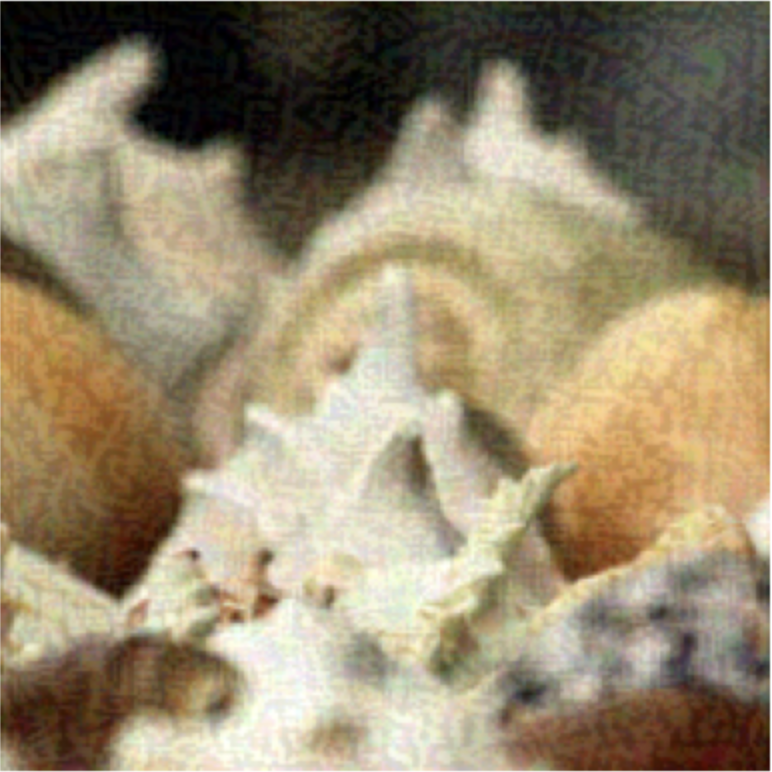} \hfill \img{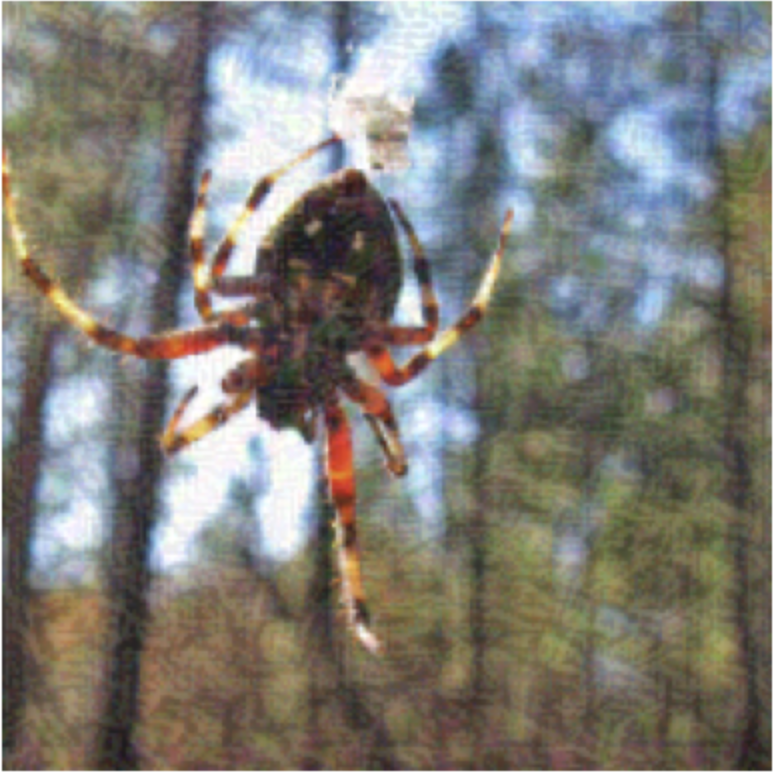} \hfill \\
  \hfill
  \begin{minipage}{1.8cm}
  \centering
      1.20 / 0.04
  \end{minipage}
  \hfill
  \begin{minipage}{1.9cm}
  \centering
      1.34 / 0.04
  \end{minipage}
  \hfill
  \begin{minipage}{1.9cm}
  \centering
      1.30 / 0.03
  \end{minipage}
  \hfill
  \begin{minipage}{1.9cm}
  \centering
      1.55 / 0.04
  \end{minipage}
  \hfill
  \begin{minipage}{1.9cm}
  \centering
      1.45 / 0.05
  \end{minipage}
  \hfill
  \begin{minipage}{1.9cm}
  \centering
      1.69 / 0.05
  \end{minipage}
  \hfill
  \begin{minipage}{1.9cm}
  \centering
      1.52 / 0.05
  \end{minipage}
  \hfill
  \caption{Adversarial examples generated using PGD with $\norm{\delta}_{\infty} \leq 0.05$ on ImageNet-100 validation set. Original and perturbed images are shown in the first and second rows, respectively. The LPIPS/R-LPIPS values for these images are mentioned below each image. In contrast with LPIPS values that are quite large, the R-LPIPS are very small and correctly reflect the small difference between images.}
  \label{fig:images}
\end{figure*}

\paragraph{Adversarial Examples.}

Since the discovery of adversarial examples~\citep{szegedy2013intriguing}, significant research has been devoted to developing attacks~\citep{goodfellow2014explaining,kurakin2018adversarial,carlini2017towards,croce2020reliable,croce2021mind} and defenses~\citep{goodfellow2014explaining,madry2017towards,pinot2019theoretical,araujo2020advocating,araujo2021lipschitz,meunier2022dynamical,araujo2023a}, resulting in an ongoing battle between the two.
Most of these defenses relied on smoothing the local neighborhood around each point, resulting in very small gradients that the attacks were based on.
However, it has become apparent that many of the proposed empirical defenses  could be circumvented with stronger attacks~\cite{athalye2018obfuscated}. 
In the context of a classification task, one of the best attacks, called Projected Gradient Descent (PGD)~\cite{madry2017towards} consists in maximizing the cross-entropy loss with respect to a perturbation added to the input and then projecting the perturbation to a specific $\ell_p$ ball.
This attack also led to one of the strongest empirical defenses~\cite{athalye2018obfuscated} called adversarial training (AT) which trains neural networks with adversarial examples crafted with PGD attack.

\paragraph{Adversarial Robustness \& Similarity Metrics.}

Perceptual similarity metrics based on deep features inherit both the emergent properties (good features) and the sensitivity to adversarial perturbation.
To the best of our knowledge, the robustness of LPIPS has only been investigated in the work proposed by~\cite{kettunen2019lpips}.
They introduced a self-ensembled metric (E-LPIPS), which operates in the space of natural images.
However, this approach may have limitations, as ensembling models have been shown to be ineffective in defending against adversarial examples~\cite{athalye2018obfuscated}.
Another line of work has proposed to use the LPIPS metric to craft a perceptual attack~\cite{laidlaw2020perceptual}.
They build upon PGD and introduced a new attack called Perceptual Projected Gradient Descent (PPGD) which consists in projecting with the LPIPS metric instead of an $\ell_p$ norm.
Furthermore, they combined an AT scheme with PPGD, called PAT, and demonstrated strong defenses against adversarial attacks that generalize to unforeseen threat models.

\section{Robust Perceptual Similarity Metric}
\label{section:adversarial_trained_lpips}

In this section, we build upon LPIPS and adversarial training and introduce R-LPIPS a new \emph{robust} perceptual similarity metric. 
Moreover, based on this new robust metric we propose two new strong perceptual attacks.

\begin{table*}[h]
  \centering
  \caption{Results of Naturally and Adversarially Trained LPIPS with respect to adversarial attacks generated using $\ell_2$ and $\ell_{\infty}$ norms against the BAPPS dataset. First, we can observe from table (a) that LPIPS is not robust to adversarial attacks crafted with $\ell_{\infty}$-PGD or $\ell_2$-PGD. We note that the $\ell_{\infty}$ attacks are causing the largest drops in 2AFC score compared to $\ell_2$-PGD attacks. Tables (b) and (c) show the 2AFC score under attack for R-LPIPS trained with $\ell_{\infty}$ and $\ell_2$. The natural 2AFC score of R-LPIPS remains mostly the same as LPIPS while the 2AFC score under attack is considerably improved. }
  \vspace{0.1cm}
  \resizebox{0.82\textwidth}{!}{%
  \begin{tabular}{llccccccccc}
  \cmidrule[1pt]{2-11}
  & & \multirow{2}{*}{\makecell{\textbf{Natural} \\ \textbf{2AFC}}} & & \multicolumn{3}{c}{\textbf{$\ell_{\infty}$-PGD ($\epsilon=$8/255)}} & & \multicolumn{3}{c}{\textbf{$\ell_2$-PGD ($\epsilon=$1.0)}}  \\
  \cmidrule{5-7} \cmidrule{9-11}
  & &  & & $x_0$  & $x_1$  & $x_0$/$x_1$ & & $x_0$ & $x_1$ & $x_0$/$x_1$ \\
  \cmidrule{2-11}
  (a) & \multicolumn{1}{l}{\textbf{LPIPS}} \\ 
  \cmidrule{2-11}
  & Traditional  & 74.58 & & 64.22 & 64.36 & 63.47 & & 71.17 & 72.92 & 69.48 \\
  & CNN-based    & 83.52 & & 70.02 & 68.37 & 68.92 & & 80.06 & 79.35 & 78.53 \\
  & Superres     & 71.36 & & 58.92 & 58.46 & 59.74 & & 65.07 & 65.54 & 63.53 \\
  & Deblur       & 60.92 & & 53.68 & 51.70 & 53.92 & & 58.15 & 57.45 & 56.55 \\
  & Color        & 65.53 & & 58.72 & 51.74 & 54.76 & & 61.97 & 57.84 & 60.07 \\
  & Frameinterp  & 63.01 & & 53.99 & 52.60 & 51.47 & & 58.89 & 58.05 & 55.01 \\
  \cmidrule{2-11}
  (b) & \multicolumn{1}{l}{\textbf{R-LPIPS with $\ell_\infty$ AT}} \\
  \cmidrule{2-11}
  & Traditional & 70.94 & & 66.29 & 66.30 & 63.89 & & 69.58 & 70.42 & 70.34 \\
  & CNN-based   & 83.04 & & 75.17 & 74.80 & 73.74 & & 81.28 & 81.68 & 80.77 \\
  & Superres    & 71.77 & & 63.28 & 61.42 & 61.34 & & 68.06 & 67.18 & 66.19 \\
  & Deblur      & 60.83 & & 54.85 & 53.58 & 57.46 & & 59.53 & 58.57 & 58.74 \\
  & Color       & 65.55 & & 57.98 & 55.56 & 59.50 & & 62.32 & 63.69 & 59.80 \\
  & Frameinterp & 63.27 & & 56.63 & 53.82 & 58.95 & & 61.38 & 56.05 & 58.46 \\
  \cmidrule{2-11}
  (c) & \multicolumn{1}{l}{\textbf{R-LPIPS with $\ell_2$ AT}} \\
  \cmidrule{2-11}
  & Traditional	& 73.19 & & 67.07 & 65.11 & 65.63 & & 71.17	& 71.35	& 72.14 \\
  & CNN	        & 83.40	& & 73.46 & 72.28 & 71.98 & & 81.56	& 81.70	& 80.27 \\
  & Superres      & 71.70	& & 60.42 & 59.89 & 58.21 & & 67.17	& 68.10	& 65.44 \\
  & Deblur	    & 61.19	& & 54.24 & 53.05 & 52.20 & & 58.69	& 57.58	& 58.58 \\
  & Color	        & 65.71	& & 59.21 & 53.83 & 57.69 & & 63.44	& 61.38	& 60.40 \\
  & Frameinterp	& 63.53	& & 53.67 & 53.56 & 58.23 & & 61.57	& 57.58	& 55.38 \\
  \cmidrule[1pt]{2-11}
  \end{tabular}
  }
 \label{tab:LPIPS_accuracies}
\end{table*}

\subsection{Adversarially Trained Perceptual Similarity Metric}
\label{subsection:R-LPIPS}

The LPIPS metric~\cite{kettunen2019lpips} is defined as the $\ell_2$ norm of deep features of a trained convolutional neural network.
More formally, for inputs $x, x_0 \in \mathcal{X}$, the LPIPS metric is defined as follows: 
\begin{equation}
  d(x, x_0) = \sum_{j}\frac{1}{W_j H_j}\sum_{h,w}\norm{\phi^j(x) - \phi^j(x_0)}_2^2
\label{eq:original_lpips}
\end{equation}
where $\phi^j(\cdot)$ is defined as:
\begin{equation} \label{eq:lpips_weights}
  \phi^j(x) = w_j \odot o_{hw}^j(x) 
\end{equation}
and $o^j(x)$ and $o^j(x_0)$ are the internal activations of a trained convolutional neural network scaled channel-wise by vector $w_j$.
Then, the $\ell_2$ norm of the weighted activations is normalized by the width and height of filters.

In order to build the LPIPS metric, \citet{zhang2018unreasonable} used the features of the AlexNet classification model trained on the ImageNet dataset~\cite{deng2009imagenet}.
Then, they ``tune'' the metric by learning the weights $w_j$ on the BAPPS dataset.
More formally, the loss used to ``tune'' the metric is defined as follows:
\begin{equation*}
  l_{\text{ce}} \left[ g_\theta \left( d_w(x, x_0), d_w(x, x_1) \right), h \right]
\end{equation*}
where $l_{\text{ce}}$ is the cross-entropy loss, $x_0$, $x_1$ are distortions of the reference images $x$ from the BAPPS dataset, $h \in (0,1)$ is a perceptual score, $g_\theta$ is a small network parameterized by $\theta$, trained to map distances to $h$ score and $d$ is the distance defined in Equation~\eqref{eq:original_lpips} and it is parameterized by $w$.

To adversarially train the LPIPS metric, we leverage the adversarial training scheme introduced by~\citet{madry2017towards} and introduce an adversarial perturbation $\delta$ at each step of the training on $x_0$: 
\begin{equation}
  \min_{\theta, w} \max_{\delta : \norm{\delta}_p \leq \varepsilon} l_{\text{ce}} \left[ g_{\theta}\left( d_w(x, x_0+\delta), d_w(x, x_1) \right), h \right]
\end{equation}
The new weights $w$ trained with adversarial training become the building block of R-LPIPS following the same construct as LPIPS in Equation~\eqref{eq:lpips_weights}.

\subsection{New Attacks based on R-LPIPS}
\label{subsection:new_attack}

Recently, LPIPS has been employed instead of $\ell_p$ norms to produce perceptual adversarial attacks.
The general constrained optimization scheme to craft an adversarial example with respect to the LPIPS metric is defined as follows:
\begin{equation*}
   \max_{\Tilde{x}} \quad l_{m}\left[ f(\Tilde{x}), y \right] \quad \text{s.t.} \quad \norm{\phi(x) - \phi(\Tilde{x})}_2 \leq \varepsilon
\end{equation*}
where $l_\text{m} = \max_{i \neq y} (f(\Tilde{x})_y - f(\Tilde{x})_i )$ is the margin loss used by~\citet{carlini2017towards}, $f(\cdot)$ is the classifier, and $\phi(\cdot)$ is the network that generates feature vectors.
However, this constrained optimization scheme is not trivial.
Therefore, \citet{laidlaw2020perceptual} relax the problem and proposed two perceptual attack methods, Perceptual Projected Gradient Descent (PPGD) and Lagrangian Perceptual Attack (LPA) based on the LPIPS metric to craft adversarial perturbations with better perceptual properties.

Perceptual Projected Gradient Descent (PPGA) tries to find the optimal $\delta $ by using first-order Taylor’s approximation and rewriting the optimization formula as:
\begin{align*}
    \max_\delta \quad l\left[ f(x), y \right] + \nabla l \left[ f(x), y) \right]^\top \delta \quad \text{s.t.} \quad \norm{J\delta}_2\leq \eta
\end{align*}
where $J$ is the Jacobian matrix of $\phi(\cdot)$ at $x$, $\delta$ is the perturbation size applied to $x$, and $\eta$ is the step size.
The second method, Lagrangian Perceptual Attack (LPA), uses a Lagrange multiplier to add the constraint to the optimization formula and perform the optimization:
\begin{equation*}
   \max_{\Tilde{x}} \quad l \left[ f(\Tilde{x}), y \right] - \lambda \max \left( 0, \norm{\phi(\Tilde{x}) -  \phi(x)}_2 - \varepsilon \right)
\end{equation*}
In this work, we build upon the work of \citet{laidlaw2020perceptual} and propose the R-PPGD and R-LPA attack scheme which consist of the same optimization but our R-LPIPS is replaced with the classical LPIPS metric. 

 % \vfill

\begin{figure*}[h!]
\centering
\hfill
\begin{subfigure}{0.43\textwidth}
  \centering
  \includegraphics[width=\textwidth]{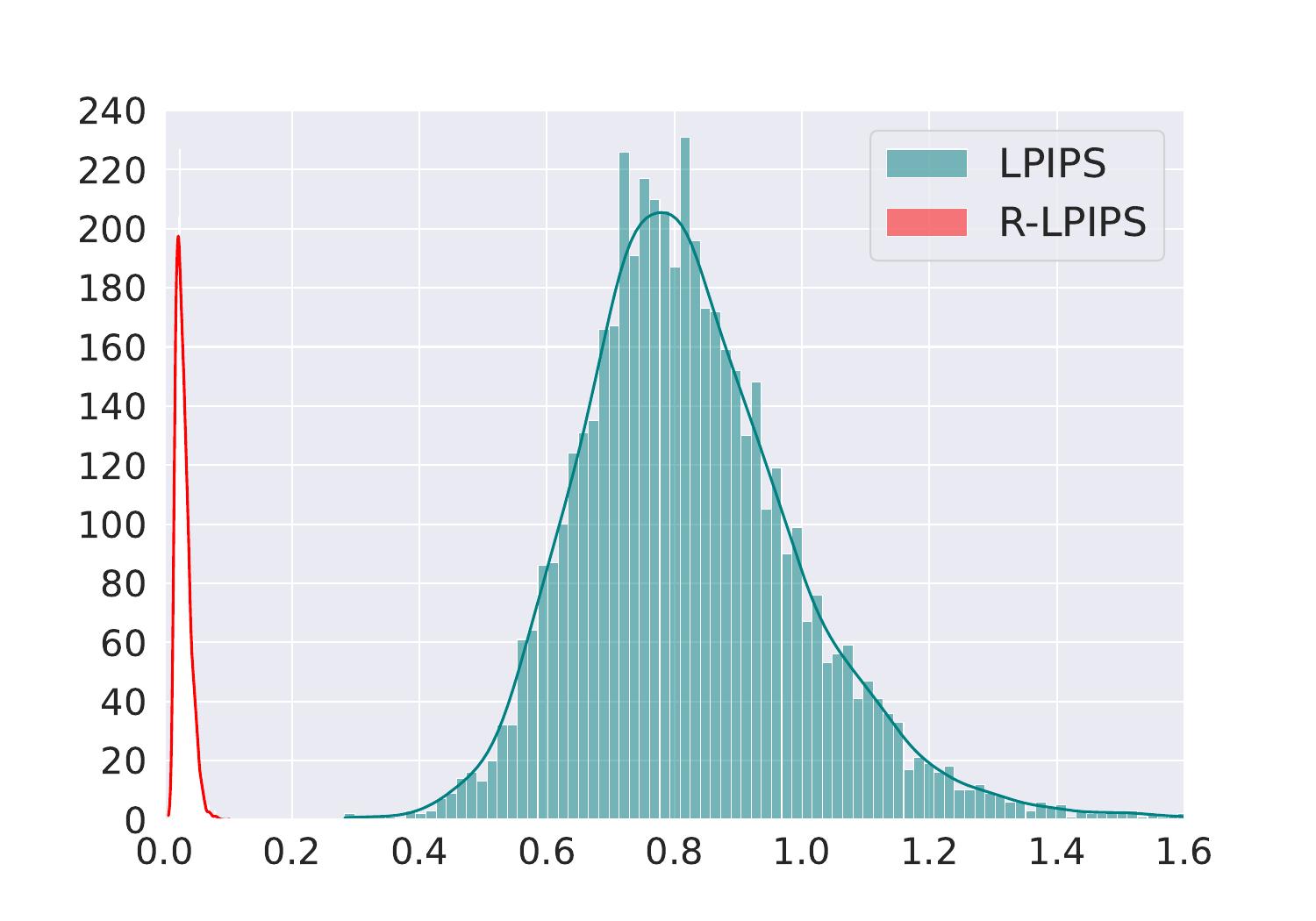}
  \caption{OPT with LPIPS}
  \label{fig:lpips_attack_histogram}
\end{subfigure}%
\hfill
\begin{subfigure}{0.43\textwidth}
  \centering
  \includegraphics[width=\textwidth]{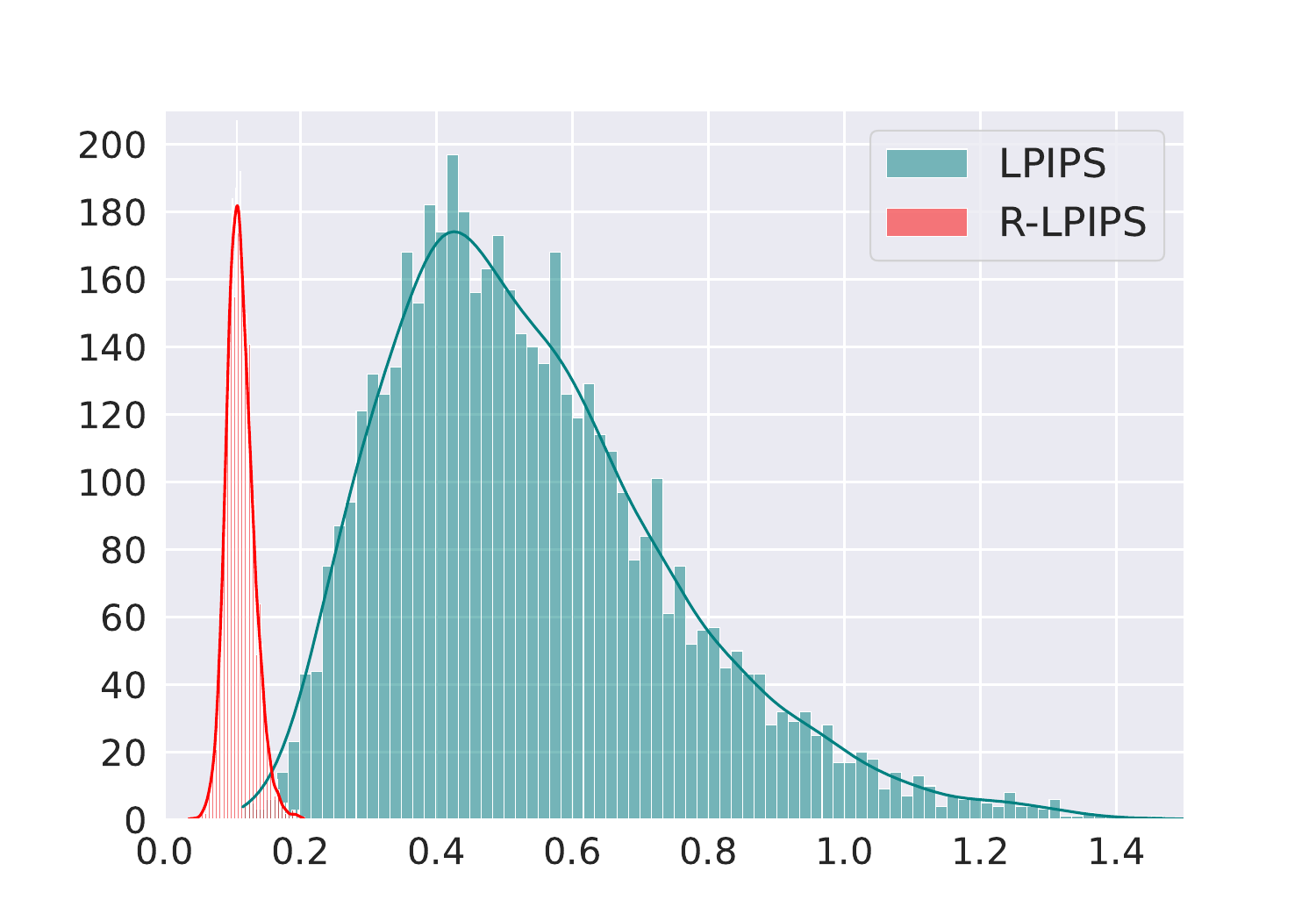}
  \caption{OPT with R-LPIPS}
  \label{fig:r_lpips_attack_histogram}
\end{subfigure}%
\hfill
\caption{The histogram of LPIPS and R-LPIPS distances between the clean and OPT adversarial examples for ImageNet-100 validation set. Figure (a) shows the LPIPS and R-LPIPS distribution of adversarial examples generated using the OPT optimization scheme and the LPIPS metric with $\norm{\delta}_{\infty} \leq 0.05$. Although LPIPS is fooled and assigns large values to the semantically identical images, the R-LPIPS shows complete robustness and considers the perturbation to be small. Figure (b) shows the same setup except R-LPIPS is used instead of LPIPS during the optimization of adversarial attack. Although R-LPIPS values in (b) are greater than the values in (a), they are still far from the threshold ($0.5$) and are quite smaller than the values of LPIPS.}
\label{fig:histogram_of_distances}
\end{figure*}

\section{Experiments}
\label{sec:experiments}
In this section, we present a comprehensive set of experiments to demonstrate the superiority of R-LPIPS compared to the classical LPIPS metric. More precisely, we aim at answering the following questions:
\begin{itemize}[parsep=-4pt,leftmargin=26pt,topsep=0pt]
    \item[\textbf{(Q1)}] Is LPIPS vulnerable to adversarial examples? 
    \item[\textbf{(Q2)}] How robust is R-LPIPS compared to LPIPS?
    \item[\textbf{(Q3)}] Can R-LPIPS leads to stronger attacks?
\end{itemize}

\subsection{Vulnerabilities of LPIPS (Q1)}

To illustrate the lack of robustness of the LPIPS metric, we present two sets of results. 
First, we present the result of $\ell_\infty$-PGD and $\ell_2$-PGD against the LPIPS metric in Table~\ref{tab:LPIPS_accuracies}a with $\varepsilon = 8/255$ and $\varepsilon = 1$ respectively on $x_0$, $x_1$ independently, and $x_0$/$x_1$ together. To evaluate the performance of LPIPS over clean and adversary data, we use the 2FAC score which was employed in \citet{zhang2018unreasonable}.
It can be observed that the 2AFC score under attack on different distortions is significantly lower, up to 15.15\% lower for $\ell_\infty$-PGD and for 8\% lower $\ell_2$-PGD, compared to the natural 2AFC score of the LPIPS metric over different distortions. 

Second, we propose a new optimization scheme, called OPT, to demonstrate that there exist adversarial examples with small $\ell_\infty$ perturbations such that the LPIPS metric is large. 
More formally, by defining $\phi(\cdot)$ as the model that generates the feature vectors, we define the optimization formula for the attack as follows:
\begin{equation*} \label{eq:adv_lpips}
   \max_{\delta : \norm{\delta}_{\infty} \leq \varepsilon} l_\text{MSE} \left[ \phi(x + \delta), \phi(x) \right]
\end{equation*}

where $l_\text{MSE}$ is the mean squared error loss and we choose $\varepsilon = 0.05$ for the optimization. 
We perform this attack on the validation set of ImageNet-100 and present the distribution of the values of LPIPS in Figure~\ref{fig:lpips_attack_histogram} in blue.
Based on the result presented by~\citet{laidlaw2020perceptual}, two images with an LPIPS value greater than 0.5 are observable to humans. 
The histogram in Figure~\ref{fig:lpips_attack_histogram} demonstrates that nearly all images of the ImageNet-100 validation set have an LPIPS value over 0.5 while having a difference of 0.05 in $\ell_\infty$ which is considered very small and nearly imperceptible to humans. 
Figure~\ref{fig:images} illustrates this difference. The top row shows clean reference images while the bottom row shows OPT adversarial images and the left value below shows the LPIPS value.

\begin{table}[t]
  \centering
  \caption{Accuracy of Perceptual Adversarial Training (PAT) variants on CIFAR-10 against PPGD/LPA and R-PPGD/R-LPA attacks with the constraint of 0.5 for perturbation size measured by LPIPS and R-LPIPS. Although PAT variants show relative robustness to PPGD/LPA attacks, they are completely vulnerable to attacks generated by R-PPGD and R-LPA.}
  \vspace{0.1cm}
  \resizebox{\columnwidth}{!}{
    \begin{tabular}{lcccccc}
    \toprule
     & & \multicolumn{1}{c}{\textbf{PPGA}} & \multicolumn{1}{c}{\textbf{LPA}} & & \multicolumn{1}{c}{\textbf{R-PPGA}} & \multicolumn{1}{c}{\textbf{R-LPA}} \\
    \midrule
    \textbf{PAT-self}    & & 13.1 & 2.1 & & 3.1 & 0.0 \\
    \textbf{PAT-AlexNet} & & 26.6 & 9.8 & & 4.3 & 0.2 \\
    \bottomrule
    \end{tabular}%
  }
  \vspace{-0.2cm}
  \label{tab:PAT_results}%
\end{table}%

\begin{figure*} [h]
  \centering
  \hfill \imgpat{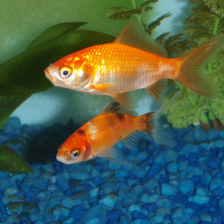} \hfill \imgpat{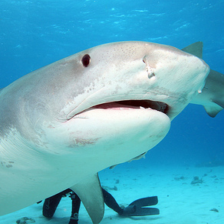} \hfill \imgpat{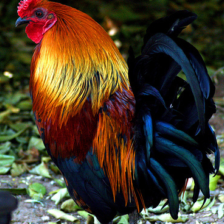} \hfill \imgpat{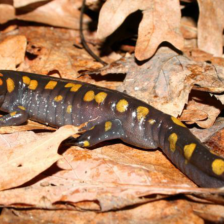} \hfill \imgpat{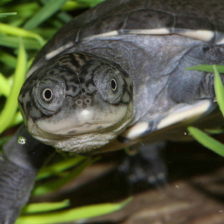} \hfill \imgpat{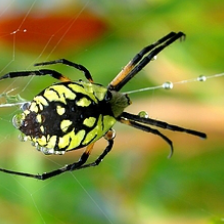} \hfill \imgpat{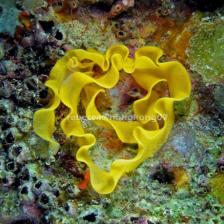} \hfill \\
  \hfill \imgpat{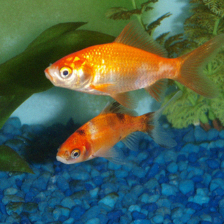} \hfill \imgpat{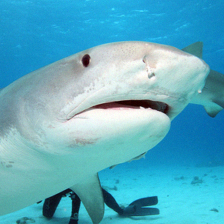} \hfill \imgpat{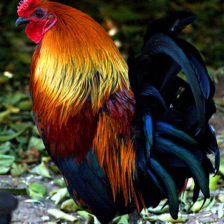} \hfill \imgpat{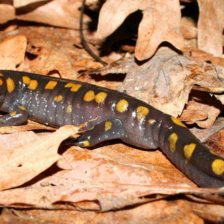} \hfill \imgpat{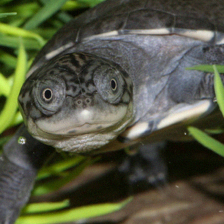} \hfill \imgpat{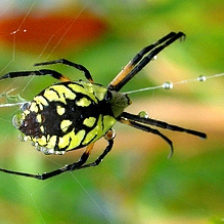} \hfill \imgpat{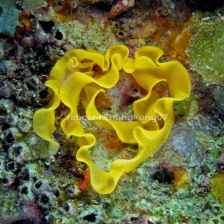} \hfill \\
  \hfill \imgpat{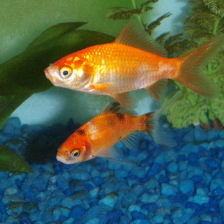} \hfill \imgpat{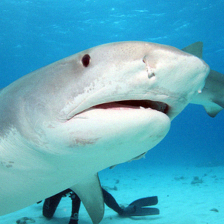} \hfill \imgpat{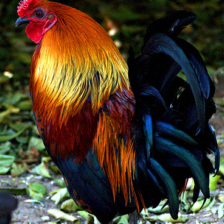} \hfill \imgpat{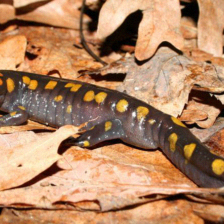} \hfill \imgpat{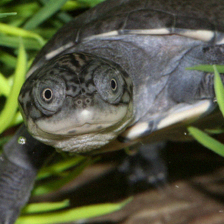} \hfill \imgpat{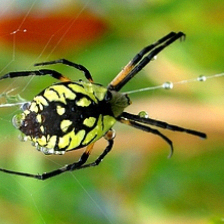} \hfill \imgpat{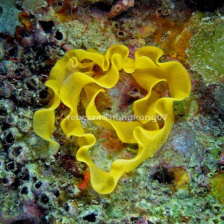} \hfill \\
  \caption{The adversarial images generated by the R-PPGA/R-LPA attacks. Original images are shown in the first row, and the adversarial images generated by R-PPGA and R-LPA (which are bounded by $0.5$) are added to the second and third rows. Similar to adversarial images generated by PPGA/LPA~\cite{laidlaw2020perceptual}, the perturbations between images in this figure are invisible to human eyes.}
  \label{fig:PPGD_LPA_images}
\end{figure*}

\subsection{Robust LPIPS (R-LPIPS) (Q2)}

After observing the vulnerabilities of LPIPS, we propose a new robust perceptual similarity metric called R-LPIPS. 
In order to develop R-LPIPS, we leverage the training scheme of LPIPS and adversarial training.
The setup is explained in detail in Section~\ref{subsection:R-LPIPS}.
First, we conducted an adversarial training with $\ell_{\infty}$ and $\ell_2$ norms and evaluated both R-LPIPS against the BAPPS dataset. 
Table~\ref{tab:LPIPS_accuracies} presents results for natural and under attack images with $\ell_\infty$-PGD and $\ell_2$-PGD against adversarial training conducted with $\ell_{\infty}$ (Table~\ref{tab:LPIPS_accuracies}b) and $\ell_2$ (Table~\ref{tab:LPIPS_accuracies}c) norms.

The first interesting result to observe from Table~\ref{tab:LPIPS_accuracies} is that the natural 2AFC score is preserved across all data distortions, except for a slight decrease observed for the traditional distortion. 
We can even observe slight improvements in the natural 2AFC score for some distortions, as the model shows a better generalization.
To evaluate robustness of R-LPIPS, we compute a perturbation with PGD attack with  $\ell_{\infty}$ and  $\ell_2$ norms with $\varepsilon = 8/255$ and $\varepsilon = 1$, respectively on $x_0$, $x_1$ independently, and $x_0$/$x_1$ together. 
We observe a consistent increase in robustness of R-LPIPS compared to the original LPIPS metric trained without AT.
We also note that R-LPIPS with $\ell_{\infty}$-AT seems to provide better results. In the following, we refer to $\ell_{\infty}$-AT LPIPS as the R-LPIPS. 

To further demonstrate the robustness of R-LPIPS with respect to LPIPS, we computed the R-LPIPS metric on the OPT adversarial examples. 
The distribution of the values of R-LPIPS is shown in Figure~\ref{fig:lpips_attack_histogram}.
One can observe that the two distributions are entirely separated and the values of R-LPIPS are very small when the $\ell_\infty$ perturbation is small. 
Figure~\ref{fig:r_lpips_attack_histogram} provides the same experiments but with OPT with R-LPIPS instead of LPIPS. The adversarial examples are therefore stronger but the distinction between LPIPS and R-LPIPS is still significant. 
To better illustrate, Figure~\ref{fig:images} provides a set of images with the comparison between the LPIPS value and the R-LPIPS value.

\subsection{Perceptual Adversarial Attack with R-LPIPS (Q3)}
% Based on the schemes for generating perceptual adversarial attacks, 
Perceptual Adversarial Training (PAT) was also proposed by~\citet{laidlaw2020perceptual} to train the model using perceptual attacks and come up with a perceptually robust model.
In this section, we combined the PPGA and LPA attacks with our robust perceptual distance metric and developed attacks named R-PPGA and R-LPA.
To compare the strength of attacks generated by LPIPS and R-LPIPS, we reproduced the accuracy of PAT to attacks generated by PPGA and LPA on CIFAR-10 and performed an experiment to compute the accuracy of PAT to R-PPGA and R-LPA attacks.
Our results (Table~\ref{tab:PAT_results}) shows that PAT has relative robustness to attacks constrained with LPIPS, and is highly vulnerable to attacks bounded by R-LPIPS.
The significant drop in the accuracy of PAT when exposed to R-PPGA and R-LPA attacks motivated us to visualize the adversarial data for attacks generated based on R-LPIPS; the results are shown in Figure~\ref{fig:PPGD_LPA_images}. The first row consists of the original images, and the second and third rows are the adversarial images generated by R-PPGA and R-LPA attacks, respectively.

\section{Conclusion \& Future Work}

\textbf{Conclusion.} In this paper, we showed that the LPIPS metric is vulnerable to adversarial examples and proposed R-LPIPS, a perceptual similarity metric that has been trained adversarially. During the process of adversarial training, the $w_l$ weights are optimized while leaving the backbone weights of the model (AlexNet architecture) unchanged.
Our findings reveal that R-LPIPS exhibits superior generalization and robustness across various data distortions when subjected to $\ell_\infty$-PGD and $\ell_2$-PGD attacks.  Additionally, we have investigated strong perceptual attacks using R-LPIPS, namely R-PPGA and R-LPA, and demonstrated their superiority over the previously established state-of-the-art attacks.

\textbf{Future work.} 
First, the R-LPIPS metric, which is an adversarially trained version of LPIPS achieved through $\ell_{\infty}$ AT on $x_0$, could be further explored by applying AT to $x_1$ or $x_0$ and $x_1$. Assessing the robustness of these different versions would offer valuable insights.
%Second, we observed that R-LPIPS demonstrates enhanced generalization capabilities on unseen data, it would be valuable to investigate this property in greater detail.
Second, R-LPIPS can be used as a defense mechanism in similar settings as the PAT training scheme and LPIPS. It would be interesting to explore the development of R-PAT, which has the potential to be a more universal perceptual adversarial defense. Evaluating its performance under attack would provide valuable insights and potentially demonstrate superior results.
Finally, by using adversarial training to develop R-LPIPS, the new metric inherits its drawback which is the lack of theoretical guarantees.
An interesting future direction would be to devise guarantees to a perceptual metric. 

\section*{Acknowledgments}
This work was supported in part by the Army Research Office under grant number W911NF-21-1-0155 and by the New York University Abu Dhabi (NYUAD) Center for Artificial Intelligence and Robotics, funded by Tamkeen under the NYUAD Research Institute Award CG010.

\bibliographystyle{icml2023}
\bibliography{bibliography}

\end{document}